%% file: my_paper.tex
\documentclass{article} 
\usepackage{iclr2026_conference,times}

\usepackage{hyperref}
\usepackage{url}

\usepackage{graphicx}
\usepackage{booktabs}
\usepackage{multirow}
\usepackage{subcaption}
\usepackage{amsmath}
\usepackage{amssymb}
\usepackage{tabularx}   
\usepackage[table]{xcolor} 
\usepackage{algorithm}
\usepackage{algorithmic}

\title{D3LM: A Discrete DNA Diffusion Language Model for Bidirectional DNA Understanding and Generation}

\author{Zhao Yang\thanks{Equal contribution.} \\
Gaoling School of Artificial Intelligence \\
Renmin University of China \\
\texttt{yangyz1230@ruc.edu.cn} \\
\And
Hengchang Liu\footnotemark[1] \\
Gaoling School of Artificial Intelligence\\
Renmin University of China \\
\texttt{liuhengchang@ruc.edu.cn} \\
\AND
Chuan Cao \\
Zhongguancun Academy \\
\texttt{chuancao@bza.edu.cn} \\
\And
Bing Su \\
Gaoling School of Artificial Intelligence\\
Renmin University of China \\
\texttt{bingsu@ruc.edu.cn} \\
}

\iclrfinalcopy 
\begin{document}

\maketitle

\begin{abstract}
Early DNA foundation models adopted BERT-style training, achieving good performance on DNA understanding tasks but lacking generative capabilities. Recent autoregressive models enable DNA generation, but employ left-to-right causal modeling that is suboptimal for DNA where regulatory relationships are inherently bidirectional. We present D3LM (\textbf{D}iscrete \textbf{D}NA \textbf{D}iffusion \textbf{L}anguage \textbf{M}odel), which unifies bidirectional representation learning and DNA generation through masked diffusion. D3LM directly adopts the Nucleotide Transformer  (NT) v2 architecture but reformulates the training objective as masked diffusion in discrete DNA space, enabling both bidirectional understanding and generation capabilities within a single model. Compared to NT v2 of the same size, D3LM achieves improved performance on understanding tasks. Notably, on regulatory element generation, D3LM achieves an SFID of 10.92, closely approaching real DNA sequences (7.85) and substantially outperforming the previous best result of 29.16 from autoregressive models. Our work suggests diffusion language models as a promising paradigm for unified DNA foundation models. We further present the first systematic study of masked diffusion models in the DNA domain, investigating practical design choices such as tokenization schemes and sampling strategies, thereby providing empirical insights and a solid foundation for future research. D3LM has been released at https://huggingface.co/collections/Hengchang-Liu/d3lm.
\end{abstract}

\input{sections/intro}
\input{sections/method}
\input{sections/experiment}
\input{sections/conclusion}
\bibliography{my_paper}
\bibliographystyle{iclr2026_conference}

\appendix
\input{sections/related_work}
\input{appendix/metrics}
\input{appendix/sampling}
\input{appendix/ablation}

\end{document}

%% file: sections/intro.tex
\section{Introduction}
\label{sec:intro}
\begin{figure*}[t]
\vskip -0.15in
  \begin{center}
    {\includegraphics[width=0.9\linewidth]{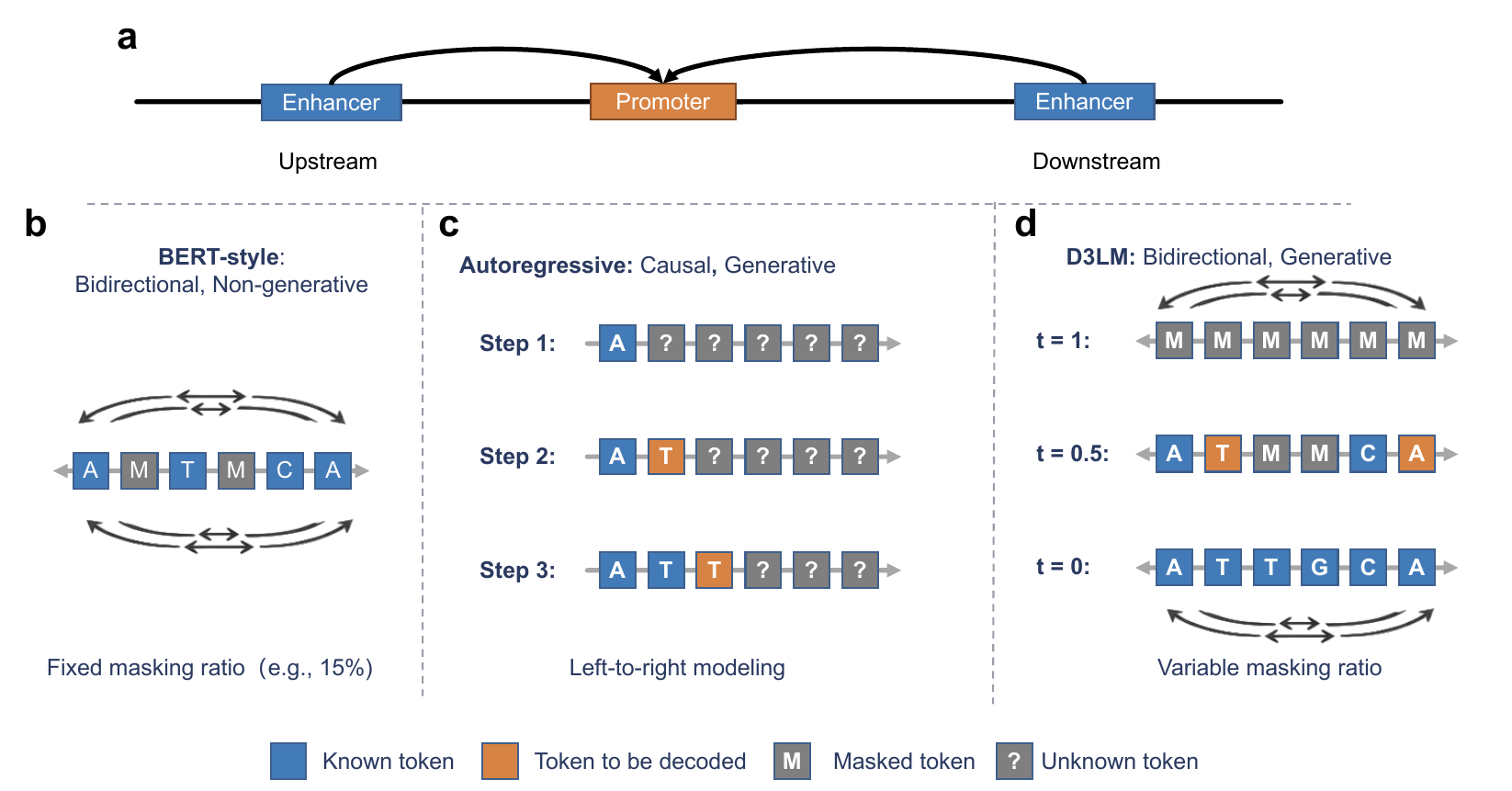}}
    \caption{\textbf{Comparison of DNA modeling approaches.} 
    \textbf{(a)} Enhancers regulate promoters from either upstream or downstream positions, demonstrating bidirectional regulatory relationships in DNA.
    \textbf{(b)} BERT-style models use bidirectional attention but employ fixed masking ratios and lack generative capabilities. 
    \textbf{(c)} Autoregressive models generate sequences left-to-right but cannot adjust earlier positions once generated, making it difficult to satisfy global constraints. 
    \textbf{(d)} D3LM combines bidirectional modeling with generation through masked diffusion with variable masking ratios, enabling iterative refinement of all positions simultaneously.}
    \label{fig:comparison}
    \label{fig:motivation}
  \end{center}
\vskip -0.2in
\end{figure*}

DNA, the blueprint of biology, encodes the fundamental instructions for life and plays a pivotal role in the central dogma of molecular biology~\citep{crick1970central}. Understanding and designing DNA sequences is crucial for applications ranging from target discovery~\citep{lindsay2003target} and personalized medicine~\citep{huang2023personal} to synthetic biology~\citep{gosai2024machine, yang2025regulatory, dasilva2025designing}. 

The recent progress of language models~\citep{devlin2019bert, brown2020language} has inspired the development of genomic foundation models that learn universal representations from large-scale unlabeled DNA sequence data. \textit{BERT-style} models such as DNABERT~\citep{DNABert, DNABert2} and Nucleotide Transformer (NT)~\citep{NT, boshar2025foundational} employ bidirectional masked language modeling, learning rich representations through predicting randomly masked nucleotides (shown in Figure~\ref{fig:motivation}(b)) While these models excel at understanding tasks and effectively capture bidirectional dependencies in DNA sequences, they lack generative capabilities. Recent works~\citep{liu2025tuna, chen2025janus} suggest that unified understanding and generation can mutually benefit model performance. On the other hand, \textit{autoregressive models} (shown in Figure~\ref{fig:motivation}(c)) such as HyenaDNA~\citep{hyenadna}, Evo~\citep{evo, merchant2026semantic}, and Generator~\citep{wu2025generator} adopt next-token prediction~\citep{brown2020language} with causal left-to-right modeling, enabling sequence generation. However, this causal modeling paradigm faces fundamental challenges when applied to DNA sequences.

Unlike natural language, where left-to-right autoregressive modeling has proven effective~\citep{kim2025train}, DNA regulatory relationships are inherently bidirectional. Multiple lines of biological evidence support this, including that transcription factor binding sites often exhibit palindromic symmetry~\citep{georgakopoulos2023transcription} and CpG islands require global GC content constraints~\citep{lal2024designing}. The most illustrative example is enhancer-promoter interactions. As shown in Figure~\ref{fig:motivation}(a), enhancer elements can regulate target genes from variable genomic positions. Some function from upstream locations~\citep{schuur1996prostate}, while others regulate from downstream~\citep{li2013far} of the promoter. Causal autoregressive models, which strictly follow left-to-right generation order, cannot adequately model downstream enhancers. These biological properties motivate the need for a DNA foundation model that combines bidirectional modeling with unified understanding and generation capabilities.


We address this challenge by introducing D3LM (\textbf{D}iscrete \textbf{D}NA \textbf{D}iffusion \textbf{L}anguage \textbf{M}odel), a unified framework that bridges understanding and generation through discrete diffusion modeling~\citep{llada}. The key distinction between D3LM and NT~\citep{NT} lies in the training objective. D3LM learns to predict masked tokens across a continuous range of masking ratios (shown in Figure~\ref{fig:motivation}(d)), while NT uses a fixed masking ratio and lacks generative capabilities. This enables D3LM to retain bidirectional modeling capabilities while gaining the ability to generate sequences. 

To isolate the impact of the training objective, we adopt the NT v2 architecture~\citep{NT}, resulting in two model variants: D3LM-R, which is randomly initialized with the NT v2 architecture, and D3LM, which is initialized from pre-trained NT v2 weights. Remarkably, D3LM demonstrates improved performance on understanding tasks compared to NT v2 while achieving strong generative capabilities that outperform state-of-the-art DNA generation models, including autoregressive models~\citep{hyenadna, evo, wu2025generator} and continuous-space latent diffusion models~\citep{discdiff}. We attribute this to the synergy between unified bidirectional understanding and generation, where both capabilities mutually reinforce each other. These results suggest that discrete diffusion language models represent a promising paradigm for DNA foundation models.

Our contributions can be summarized as follows:
\begin{itemize}
    \item We propose D3LM, a unified DNA foundation model that combines bidirectional modeling with generative capabilities through masked diffusion in discrete DNA space.

    \item D3LM demonstrates stronger representation learning than NT v2 of comparable size on downstream understanding tasks, indicating that the masked diffusion generative objective does not degrade representational quality and may even improve it.
    
    
    \item On regulatory element generation, D3LM achieves an SFID of 10.92, closely approaching real biological sequences (7.85) and substantially outperforming autoregressive models (29.16) and continuous-space latent diffusion approaches (62.74), demonstrating the advantages of bidirectional generative modeling directly in the discrete sequence space.

    \item We present the first systematic analysis of masked diffusion models applied to DNA, focusing on practical design aspects and delivering empirical insights that facilitate future work.
    
\end{itemize}

%% file: sections/method.tex
\section{Method}

In this section, we first introduce the bidirectional probabilistic formulation based on discrete masked diffusion models in Section~\ref{sec:formulation}. We then present the details of our generative training approach in Section~\ref{sec:pretraining}, including model architecture and tokenization strategy. In Section~\ref{sec:sampling}, we describe the sampling algorithm for sequence generation with D3LM. Finally, in Section~\ref{sec:finetuning}, we explain how D3LM can serve as a representation extractor for fine-tuning on diverse genomic downstream tasks.

\subsection{Bidirectional Probabilistic Formulation}
\label{sec:formulation}

We consider a DNA sequence $\mathbf{x} = (x^1, x^2, \ldots, x^L)$, where each $x^i \in \{A, C, G, T\}$ and $L$ is the sequence length. The sequence is drawn from the true distribution $\mathbf{x} \sim q(\mathbf{x})$, where $q$ represents the real DNA sequence distribution. Our goal is to learn this true distribution through generative modeling using a parametric model $p_{\theta}(\mathbf{x})$.

\textbf{Autoregressive DNA models.} 
Mainstream generative DNA language models such as HyenaDNA~\citep{hyenadna} and Evo~\citep{evo, merchant2026semantic} employ autoregressive modeling:
\begin{equation}
\underbrace{p_{\theta}(\mathbf{x}) = p_{\theta}(x^1) \prod_{i=2}^{L} p_{\theta}(x^i | x^1, \ldots, x^{i-1})}_{\text{Autoregressive formulation}}.
\label{eq:autoregressive}
\end{equation}
While this formulation is well-suited for natural language processing~\citep{kim2025train} due to the inherent left-to-right structure of language, DNA sequences do not necessarily follow such sequential dependencies as discussed in Section~\ref{sec:intro}.

\textbf{Non-generative masked models.}
Another class of DNA foundation models, such as NT~\citep{NT} and DNABERT~\citep{DNABert, DNABert2}, adopts a non-generative masked prediction strategy inspired by BERT~\citep{devlin2019bert}. These models randomly mask a fixed ratio (typically 15\%) of positions in the input sequence and train the model to predict the masked tokens using bidirectional attention. While this approach effectively captures bidirectional dependencies in DNA sequences, it lacks generative capabilities because the fixed masking ratio prevents the training objective from being a proper probabilistic generative model. Recent work~\citep{liu2025tuna, chen2025janus} suggests that unified understanding and generation can mutually benefit model performance, motivating the development of models that combine both capabilities.

\begin{figure}[t]
\vskip -0.1in
  \begin{center}
    \centerline{\includegraphics[width=0.65\linewidth]{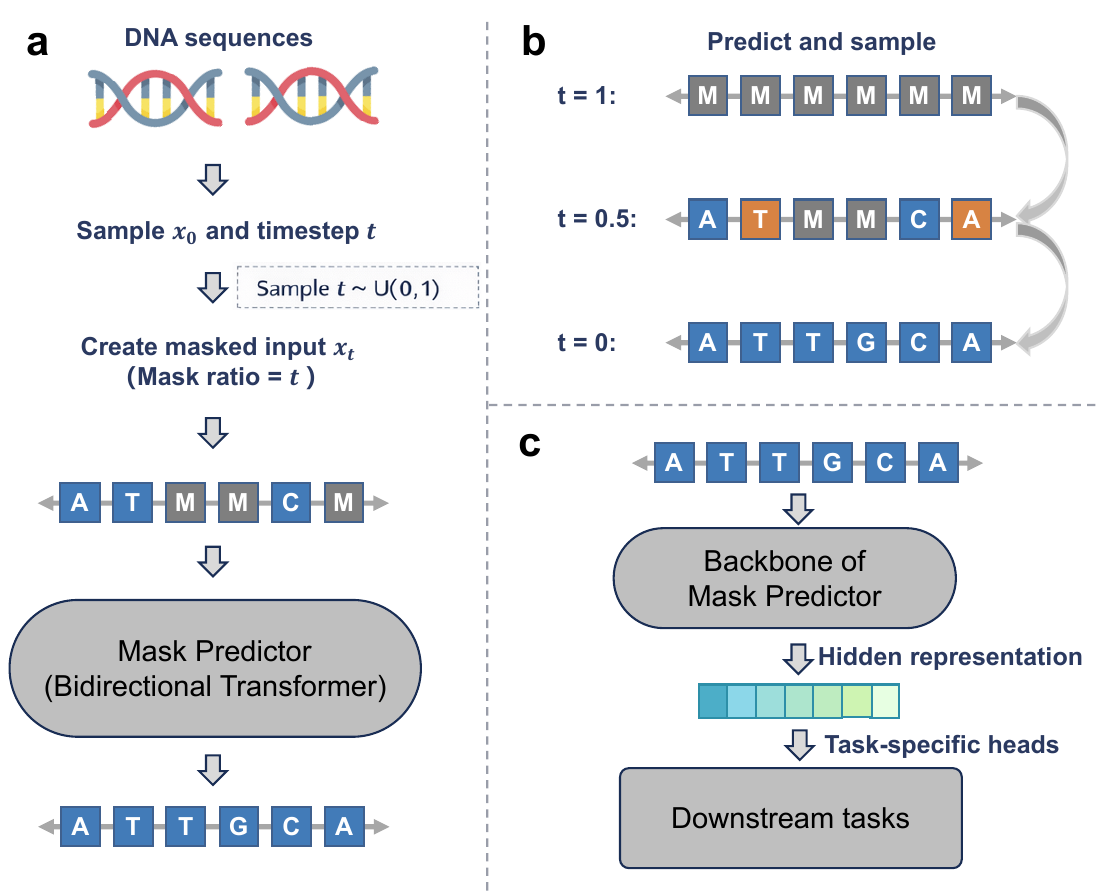}}
    \caption{\textbf{D3LM framework overview.} 
    \textbf{(a)} Training with variable masking ratios sampled from a uniform distribution. The model learns to predict masked tokens using bidirectional attention.
    \textbf{(b)} Iterative generation starting from fully masked sequences, progressively unmasking tokens through repeated sampling.
    \textbf{(c)} Fine-tuning on downstream genomic tasks using frozen or trainable encoder with task-specific heads for promoter classification, histone modification, and splice site prediction.}
    \label{fig:method}
  \end{center}
\vskip -0.2in
\end{figure}

\textbf{Bidirectional discrete DNA diffusion language models}.
To combine the strengths of both paradigms, we adopt an alternative modeling approach following discrete diffusion models~\citep{llada, dream}. Unlike the autoregressive formulation in Eq.~\eqref{eq:autoregressive}, D3LM defines a generative model distribution $p_{\theta}(\mathbf{x}_0)$ through a \textit{forward masking process} and a \textit{reverse denoising process}, following the masked diffusion language model framework following ~\citet{llada}.

The forward process gradually masks tokens independently in $\mathbf{x}_0$ until the sequence is fully masked at $t = 1$. For $t \in (0, 1)$, the sequence $\mathbf{x}_t$ is partially masked, with each position being masked with probability $t$ and remaining unmasked with probability $1 - t$. The reverse process recovers the data distribution by iteratively predicting masked tokens as $t$ moves from 1 to 0.

Similar to NT~\citep{NT}, D3LM employs a \textit{mask predictor}, a parametric model $p_{\theta}(\cdot | \mathbf{x}_t)$ that takes $\mathbf{x}_t$ as input and predicts all masked tokens simultaneously using bidirectional attention. To isolate the impact of the discrete diffusion training objective, D3LM directly adopts the NT v2 architecture~\citep{NT}, allowing us to attribute performance differences specifically to the training paradigm rather than architectural innovations. The model is trained using a cross-entropy loss computed only on the masked tokens:
\begin{equation}
\mathcal{L}(\theta) = -\mathbb{E}_{t, \mathbf{x}_0, \mathbf{x}_t} \left[ \frac{1}{t} \sum_{i=1}^{L} \mathbf{1}[x_t^i = \text{[M]}] \log p_{\theta}(x_0^i | \mathbf{x}_t) \right],
\label{eq:loss}
\end{equation}
where $\mathbf{x}_0$ denotes a clean sample from the DNA training dataset, $t$ is sampled uniformly from $[0, 1]$, and $\mathbf{x}_t$ is sampled from the forward process defined above. The loss computation is restricted to masked tokens only, as enforced by the indicator function $\mathbf{1}[\cdot]$.

Once trained, D3LM can simulate a reverse process parameterized by the mask predictor and define the model distribution $p_{\theta}(\mathbf{x}_0)$ as the marginal distribution induced at $t = 0$. The loss in Eq.~\eqref{eq:loss} provides an upper bound on the negative log-likelihood~\citep{sahoo2024simple, ou2024your}:
\begin{equation}
-\mathbb{E}_{\mathbf{x}_0 \sim q(\mathbf{x})} \left[\log p_\theta(\mathbf{x}_0)\right] \leq \mathcal{L}(\theta),
\label{eq:elbo}
\end{equation}
making it a principled objective for generative modeling. This masked modeling approach offers several advantages for DNA sequences, including the ability to capture bidirectional dependencies and generate sequences in a non-sequential manner.

\subsection{Details of Generative Training}
\label{sec:pretraining}

Previous BERT-style and autoregressive DNA models have extensively studied various design choices, including model architecture~\citep{NT, Caduceus, hyenadna}, model scale~\citep{NT, evo}, and tokenization strategies~\citep{DNABert2, wu2025generator}. Recognizing the importance of these choices, we also conduct systematic studies on key components, with detailed ablation results presented in Section~\ref{sec:ablation}. Here we present our final design choices.

\textbf{Model Architecture.}
D3LM employs the NTv2~\citep{NT} as the backbone for the mask predictor, a Transformer-based~\citep{vaswani2017attention} architecture that has proven successful for DNA representation learning. Compared to the original NTv1~\citep{NT}, NTv2 introduces key architectural improvements including Rotary Position Embeddings (RoPE)~\citep{su2024roformer} and SwiGLU activations~\citep{shazeer2020glu}. We utilize bidirectional attention to enable bidirectional modeling, which fundamentally differs from GPT-style autoregressive DNA generative models~\citep{hyenadna, evo, wu2025generator}. We experiment with four model sizes ranging from 50M to 500M parameters, following the scaling configurations of NT v2~\citep{NT}.

\textbf{Tokenization.}
Tokenization strategy is critical for genomic language models~\citep{wu2025generator}. We experiment with four different tokenization schemes: 1-mer (single nucleotide), 3-mer, 6-mer, and 9-mer. Following~\citet{NT} and~\citet{wu2025generator}, we adopt a non-overlapping 6-mer tokenization strategy, which achieves the best balance between vocabulary size and sequence representation. This results in a vocabulary size of 4105, comprising 4096 possible 6-mers ($4^6 = 4096$) plus 9 special tokens including the mask token [M], padding token [PAD], and other control tokens.

\subsection{Sampling}
\label{sec:sampling}

Once trained, D3LM generates novel DNA sequences by simulating the reverse denoising process. Starting from a fully masked sequence, we iteratively predict the clean sequence $x_0$ and selectively unmask tokens until all positions are revealed.

\textbf{Discrete-time sampling.}
We discretize the denoising process into $T$ steps. At each step $t$, the model takes the current partially masked sequence $\mathbf{x}_t$ as input and predicts the probability distribution $p_{\theta}(x_0^i | \mathbf{x}_t)$ over the original clean tokens for all masked positions. Let $\mathcal{M}_t$ denote the set of currently masked indices. Based on these predictions, we select a subset $\mathcal{U}_t \subset \mathcal{M}_t$ to unmask for the next step. The next state $\mathbf{x}_{t-1}$ is constructed via:
\begin{equation}
\mathbf{x}_{t-1}^i = \begin{cases}
\text{sample from } p_{\theta}(x_0^i | \mathbf{x}_t) & \text{if } i \in \mathcal{U}_t \\
\text{[M]} & \text{if } i \in \mathcal{M}_t \setminus \mathcal{U}_t \\
x_t^i & \text{otherwise}
\end{cases}
\label{eq:sampling_step}
\end{equation}

\textbf{Token selection strategy.}
We experimented with multiple strategies for selecting which positions to unmask, including confidence-based approaches (MaskGit~\citep{chang2022maskgit}, entropy-based sampling, top-k margin, and re-masking mechanisms). Surprisingly, we find that \textbf{random sampling}—where positions $\mathcal{U}_t$ are selected uniformly at random from $\mathcal{M}_t$—achieves the best generation quality (Table~\ref{tab:sampling}). This aligns with the training objective where masks are placed uniformly, and may reflect the non-local nature of DNA regulatory dependencies where confidence scores do not reliably indicate optimal generation order. The number of positions to unmask at each step is determined by a predefined schedule (e.g., linear or cosine). Detailed comparisons of all strategies are provided in Appendix~\ref{appendix:sampling}.

\textbf{Temperature scaling.}
To control generation diversity, we adjust the sampling distribution via temperature scaling:
\begin{equation}
p_{\theta}^{\text{temp}}(x_0^i | \mathbf{x}_t) \propto p_{\theta}(x_0^i | \mathbf{x}_t)^{1/\tau},
\label{eq:temperature}
\end{equation}
where $\tau > 0$ is the temperature parameter. Lower temperatures ($\tau < 1$) produce deterministic sequences closely following the training distribution, while higher temperatures ($\tau > 1$) increase diversity at the cost of biological fidelity. We use $\tau=1.1$ and $T=50$ steps by default (ablations in Section~\ref{sec:ablation}).

\subsection{Representation learning}
\label{sec:finetuning}
D3LM is trained to predict masked tokens at all masking levels, including the original unmasked data (i.e., masking ratio at 0\%). As a result, D3LM can simultaneously serve as a DNA sequence representation learner over massive genomic data, providing useful sequence embeddings for various downstream predictive tasks, such as sequence-level or position-level classification and regression. The sequence embedding can be obtained by simply letting D3LM take as input the given DNA sequence: $\mathbf{h}(\mathbf{x}) = \text{D3LM}_{\theta}(\mathbf{x}, t=0) \in \mathbb{R}^{L \times d}$, where $d$ is the embedding dimension and $L$ is the sequence length.

\textbf{Task-specific adaptation.}
For a downstream task with labeled data $\{(\mathbf{x}_n, y_n)\}_{n=1}^N$, where $y_n$ is the task-specific label, we add a lightweight task-specific head on top of the pre-trained encoder. For sequence-level tasks~\citep{NT} (e.g., promoter classification), we aggregate token representations using mean pooling: $\bar{\mathbf{h}} = \frac{1}{L}\sum_{i=1}^{L} \mathbf{h}_i(\mathbf{x})$. For position-level tasks, we directly use the token-level representations $\mathbf{h}_i(\mathbf{x})$. The model is then fine-tuned using task-specific objectives:
\begin{equation}
\mathcal{L}_{\text{task}}(\theta, \phi) = -\sum_{n=1}^{N} \log p_{\phi}(y_n | \mathbf{h}(\mathbf{x}_n)),
\label{eq:finetune}
\end{equation}
where $\phi$ denotes the parameters of the task-specific head. 

%% file: sections/experiment.tex
\section{Experiment}

We evaluate D3LM on extensive understanding and generative tasks, starting with the generative training details in Section~\ref{sec:training_details}. We assess DNA generation capabilities in Section~\ref{sec:unconditional} and performance on predictive downstream tasks in Section~\ref{sec:nt_downstream}. Finally, we conduct ablation studies on tokenization strategies, model scale and sampling methods in Section~\ref{sec:ablation}.

\subsection{Generative Training Details}
\label{sec:training_details}

\textbf{Dataset and Tokenization.} We trained D3LM on the \textbf{EPD-GenDNA} dataset~\citep{discdiff}, a multi-species genomic corpus comprising 160k DNA sequences across 15 species, each with a fixed length of 2048 base pairs (bp). To focus on mammalian genomics, we curated a subset restricted to mammalian sequences~\citep{discdiff}, which constitute approximately 50\% of the total corpus ($\approx$ 80,000 sequences). Adopting the tokenization strategy of the NT~\citep{NT}, we process DNA sequences using non-overlapping \textbf{6-mers}. This yields a vocabulary size of 4,105, comprising all possible 6-mers ($4^6 = 4,096$) and 9 special tokens.

\textbf{Optimization and Initialization.} The model was optimized by minimizing the discrete diffusion objective defined in Eq.~\eqref{eq:loss}. We employed the AdamW optimizer with $\beta_1 = 0.9$ and $\beta_2 = 0.95$. The learning rate followed a cosine decay schedule, peaking at $8 \times 10^{-5}$ after a linear warmup phase of the first 5\% of training steps. To ensure training stability, we applied a weight decay of 0.01 and a gradient clipping threshold of 1.0. The effective batch size is set to 32 sequences, corresponding to approximately 65,536 \textbf{base pairs} per step. We consider two model variants: \textbf{D3LM-R}, initialized randomly with the NT v2 architecture, and \textbf{D3LM}, initialized from pre-trained NT v2 weights.

\textbf{Computational Infrastructure.} Generative training is performed on a cluster of NVIDIA A800 GPUs with Mixed Precision (BF16) training for computational efficiency. The resulting models are evaluated on both generative and discriminative tasks, with detailed performance metrics provided in Section~\ref{sec:unconditional} and Section~\ref{sec:nt_downstream}.

\subsection{Performance of Unconditional Generation}
\label{sec:unconditional}

\textbf{Evaluation Setup and Baselines.}
We evaluated the generative performance on the EPD-GenDNA dataset~\citep{discdiff, li2024absorb} by conducting \textit{de novo} generation of 2048bp DNA sequences. The dataset was partitioned into training, validation, and test splits with an 8:1:1 ratio, using the latter solely for subsequent evaluation. To benchmark D3LM against distinct modeling paradigms, we compare it with three classes of baselines: (1) \textbf{Autoregressive (AR) models}, including HyenaDNA~\citep{hyenadna}and Evo~\citep{evo}, which rely on causal left-to-right modeling; (2) \textbf{Latent Diffusion}, represented by DiscDiff~\citep{discdiff, li2024absorb}, which projects DNA into continuous space; and (3) \textbf{Discrete Diffusion}. We adapt the protein-specific DPLM (150M)~\citep{dplm} to genomics. Key differences from our approach include its use of single-nucleotide (1-mer) tokenization and a linear $(1-t)$ loss weighting, rather than our $1/t$ inverse weighting. We also include real validation sequences (Truth) and uniform random sampling (Random) as reference bounds. 


Quality is assessed by comparing 1,000 generated sequences to a held-out reference set using four complementary metrics. We report Diversity (Div) and Novelty (Nov) to verify that the model generates varied sequences without memorizing the training corpus. Div is the average pairwise Levenshtein distance among generated sequences, and Nov is the average Levenshtein distance from each generated sequence to its nearest neighbor in the training set. For compositional validity, we use the G/C ratio (G count divided by C count), which reflects strand symmetry under Chargaff’s parity rule (typically $\approx 1.0$ in natural DNA). Finally, we measure functional fidelity with Sei-based Fréchet Inception Distance (SFID)~\citep{discdiff}: sequences are embedded by Sei~\citep{chen2022sequence} into a regulatory feature space, and SFID computes the distributional distance between generated and real sequences in that space (lower is better). Metric details are provided in Appendix~\ref{appendix:metrics}.

\textbf{Results.}
As detailed in Table~\ref{tab:uncond_gen}, D3LM-R achieves superior generation fidelity. While most models exhibit comparable Diversity and Novelty—indicating no trivial memorization—significant performance distinctions emerge in biological plausibility measures. D3LM-R attains an SFID of 10.92, closely approximating the real data baseline (Truth: 7.85) and substantially outperforming both continuous latent diffusion (DiscDiff: 62.74) and autoregressive baselines (HyenaDNA: 29.16). Notably, Evo suffers from severe distributional mismatch (SFID $>500$) and distorted GC ratio (0.86). We attribute this to the challenge of adapting large-scale pretrained models to our specific dataset: these models were trained on substantially larger and more diverse genomic corpora, and may require careful fine-tuning or domain adaptation to generalize to the regulatory element generation task. In contrast, D3LM-R preserves a GC ratio (1.07) nearly identical to natural sequences (1.06), suggesting that our discrete diffusion framework effectively captures global genomic constraints that other paradigms miss.


\begin{table*}[h]
  \caption{Evaluation results of unconditional generation on different sequence lengths.}
  \vskip -0.1in
  \label{tab:uncond_gen}
  \begin{center}
    \begin{small}
      \begin{sc}
        \begin{tabular}{lcccccccc}
          \toprule
           & \multicolumn{4}{c}{Sequence Length = 256} & \multicolumn{4}{c}{Sequence Length = 2048} \\
           \cmidrule(lr){2-5} \cmidrule(lr){6-9}
          Model & Div ($\uparrow$) & Nov ($\uparrow$) & GC & SFID ($\downarrow$) & Div ($\uparrow$) & Nov ($\uparrow$) & GC & SFID ($\downarrow$) \\
          \midrule
          Truth & 189.0 & 170.0 & 1.10& 31.94 & 1529& 1470& 1.06& 7.85 \\
          Random & 192.0 & 176.0 & 1.02& 148.20 & 1537& 1491& 1.00& 122.97\\
          \midrule
          HyenaDNA & \textbf{192.0} & \textbf{176.0} & 1.05& 122.79 & \textbf{1533}& 1479& 1.05& 29.16\\
          Evo & 191.0 & 170.0 & 1.51& 123.06 & \textbf{1533}& 1471& 0.86& 1359.98 \\
          DPLM & 191.0 & 175.0 & 0.91& 124.96 & 1531& \textbf{1485}& 0.85& 95.34\\
          DiscDiff & 191.0 & \textbf{176.0} & 0.96& 147.90 & 1527& 1482& 1.03& 62.74\\
          \midrule
          D3LM & 191.0& 172.0& 1.75& 85.41& 1531& 1474& 1.10& 25.21\\
          D3LM-R & \textbf{192.0} & 173.0 & 1.25& \textbf{74.14} & 1528& 1471& 1.07& \textbf{10.92}\\
          \bottomrule
        \end{tabular}
      \end{sc}
    \end{small}
  \end{center}
  \vskip -0.2in
\end{table*}

\subsection{Downstream Classification Tasks}
\label{sec:nt_downstream}

\textbf{Datasets.}
We evaluated representation quality on the NT downstream benchmark suite~\citep{NT}, including histone modification prediction, enhancer and promoter classification, and splice site prediction. All tasks are based on the revised NT downstream dataset released on HuggingFace. We follow the same dataset splits and evaluation pipeline as NT.

\textbf{Baselines.}
We compare D3LM against DNABERT-2~\citep{DNABert}, Enformer~\citep{enformer}, HyenaDNA~\citep{hyenadna}, and NT-MSv2 (50M)~\mbox{\citep{NT}} under the same dataset and evaluation protocol. Our methods include D3LM (50M), initialized from NT-MSv2 weights and further pre-trained with our discrete diffusion objective, and D3LM-R (50M), which uses the same architecture but starts from random initialization before discrete pre-training.

\textbf{Fine-tuning details.}
We finetuned all models using the NT training pipeline implemented with HuggingFace \texttt{Trainer} and LoRA~\citep{hu2022lora}. LoRA is applied to the attention projection modules (\texttt{query} and \texttt{value}) with rank $r{=}1$, $\alpha{=}32$, and dropout $0.1$. We trained for up to 1,000 update steps with a cosine learning-rate schedule and early stopping (patience 4), selecting the checkpoint that maximizes validation MCC. The final test performance is also reported using MCC, with detailed metric definitions provided in Appendix~\ref{appendix:metrics}.

\textbf{Results and analysis.}
Table~\ref{nt-task-table} reports MCC results. D3LM consistently matches or improves upon NT-MSv2, with the most pronounced gains on splice site prediction: D3LM achieves 0.947/0.945/0.959 on splice acceptor/site/donor, exceeding NT-MSv2 (0.922/0.928/0.915) and substantially outperforming DNABERT-2 and Enformer. This improvement is expected, as NT's masked language modeling objective (fixed 15\% masking) can be viewed as a special case of our variable-ratio masked diffusion training. We hypothesize that training on NT's original large-scale genomic corpus with our diffusion objective would yield even stronger performance. In contrast, D3LM-R shows substantially lower performance (e.g., splice site: 0.609 vs. 0.945), which we attribute to the limited scale of our training data (80K sequences) being insufficient to learn robust genomic representations from scratch without leveraging pretrained knowledge.

\begin{table*}[t]
\vskip -0.1in
  \caption{Performance on downstream NT understanding tasks (MCC $\pm$ Std).}
  \vskip -0.1in
  \label{nt-task-table}
  \begin{center}
    \begin{small}
      \begin{sc}
      \resizebox{\textwidth}{!}{
        \begin{tabular}{lcccccc}
        \toprule
        \textbf{Task} & \textbf{DNABERT-2} & \textbf{Enformer} & \textbf{HyenaDNA} & \textbf{NT-MSv2 (50M)} & \textbf{D3LM (50M)} & \textbf{D3LM-r (50M)} \\
        \midrule
        H2AFZ & 0.49 ($\pm$0.013) & \textbf{0.522 ($\pm$0.019)} & 0.467 ($\pm$0.012) & 0.501 ($\pm$0.012) & 0.501 ($\pm$0.020) & 0.451 ($\pm$0.017) \\
        H3K27ac & 0.491 ($\pm$0.01) & \textbf{0.52 ($\pm$0.015)} & 0.421 ($\pm$0.01) & 0.483 ($\pm$0.012) & 0.497 ($\pm$0.019) & 0.381 ($\pm$0.017) \\
        H3K27me3 & \textbf{0.599 ($\pm$0.01)} & 0.552 ($\pm$0.007) & 0.55 ($\pm$0.009) & 0.595 ($\pm$0.004) & 0.590 ($\pm$0.029) & 0.547 ($\pm$0.014) \\
        H3K36me3 & \textbf{0.637 ($\pm$0.007)} & 0.567 ($\pm$0.017) & 0.553 ($\pm$0.011) & 0.607 ($\pm$0.006) & 0.602 ($\pm$0.019) & 0.524 ($\pm$0.009) \\
        H3K4me1 & 0.49 ($\pm$0.008) & \textbf{0.504 ($\pm$0.021)} & 0.423 ($\pm$0.016) & 0.478 ($\pm$0.012) & 0.479 ($\pm$0.016) & 0.402 ($\pm$0.020) \\
        H3K4me2 & 0.558 ($\pm$0.013) & \textbf{0.626 ($\pm$0.015)} & 0.515 ($\pm$0.018) & 0.549 ($\pm$0.014) & 0.537 ($\pm$0.024) & 0.488 ($\pm$0.014) \\
        H3K4me3 & \textbf{0.646 ($\pm$0.008)} & 0.635 ($\pm$0.019) & 0.603 ($\pm$0.02) & 0.627 ($\pm$0.013) & 0.634 ($\pm$0.028) & 0.597 ($\pm$0.025) \\
        H3K9ac & 0.564 ($\pm$0.013) & \textbf{0.593 ($\pm$0.02)} & 0.487 ($\pm$0.025) & 0.538 ($\pm$0.011) & 0.508 ($\pm$0.026) & 0.447 ($\pm$0.026) \\
        H3K9me3 & 0.443 ($\pm$0.025) & 0.453 ($\pm$0.016) & 0.419 ($\pm$0.03) & 0.43 ($\pm$0.018) & \textbf{0.464 ($\pm$0.039)} & 0.315 ($\pm$0.033) \\
        H4K20me1 & \textbf{0.655 ($\pm$0.011)} & 0.606 ($\pm$0.016) & 0.59 ($\pm$0.007) & 0.637 ($\pm$0.008) & 0.641 ($\pm$0.020) & 0.574 ($\pm$0.020) \\
        Enhancer & 0.517 ($\pm$0.011) & \textbf{0.614 ($\pm$0.01)} & 0.476 ($\pm$0.021) & 0.511 ($\pm$0.006) & 0.556 ($\pm$0.023) & 0.431 ($\pm$0.046) \\
        Enhancer types & 0.476 ($\pm$0.009) & \textbf{0.573 ($\pm$0.013)} & 0.445 ($\pm$0.009) & 0.459 ($\pm$0.006) & 0.517 ($\pm$0.011) & 0.403 ($\pm$0.018) \\
        Promoter all & 0.754 ($\pm$0.009) & 0.745 ($\pm$0.012) & 0.698 ($\pm$0.011) & 0.731 ($\pm$0.01) & \textbf{0.764 ($\pm$0.019)} & 0.722 ($\pm$0.015) \\
        Promoter n-TATA & \textbf{0.769 ($\pm$0.009)} & 0.763 ($\pm$0.012) & 0.729 ($\pm$0.009) & 0.756 ($\pm$0.01) & 0.761 ($\pm$0.018) & 0.745 ($\pm$0.014) \\
        Promoter TATA & 0.784 ($\pm$0.036) & 0.793 ($\pm$0.026) & 0.666 ($\pm$0.041) & 0.747 ($\pm$0.043) & \textbf{0.867 ($\pm$0.081)} & 0.765 ($\pm$0.042) \\
        Splice acceptor & 0.837 ($\pm$0.006) & 0.749 ($\pm$0.007) & 0.808 ($\pm$0.009) & 0.922 ($\pm$0.006) & \textbf{0.947 ($\pm$0.012)} & 0.711 ($\pm$0.090) \\
        Splice site all & 0.855 ($\pm$0.005) & 0.739 ($\pm$0.011) & 0.907 ($\pm$0.018) & 0.928 ($\pm$0.005) & \textbf{0.945 ($\pm$0.013)} & 0.609 ($\pm$0.057) \\
        Splice donor & 0.861 ($\pm$0.004) & 0.78 ($\pm$0.007) & 0.915 ($\pm$0.047) & 0.915 ($\pm$0.006) & \textbf{0.959 ($\pm$0.006)} & 0.798 ($\pm$0.021) \\
        \bottomrule
        \end{tabular}
        }
      \end{sc}
    \end{small}
  \end{center}
\vskip -0.2in
\end{table*}

\subsection{Ablation}
\label{sec:ablation}
In this section, we present ablation studies on tokenization strategies, model scaling, sampling algorithms, denoising steps, and temperature scaling for unconditional generation. Unless otherwise stated, we use the D3LM-R model to generate 2048bp sequences and evaluate them on the EPD-GenDNA dataset, maintaining training and evaluation protocols identical to those in Section~\ref{sec:unconditional}.

\textbf{Tokenization.}
As shown in Figure~\ref{fig:tokenization}, 6-mer tokenization yields the best overall fidelity (SFID 10.92), outperforming 1-mer (15.77), 3-mer (29.31), and 9-mer (32.01). This suggests that 6-mers provide a favorable inductive bias: they are expressive enough to capture local genomic motifs and short-range dependencies, yet avoid the severe sparsity and rare-transition issues introduced by larger k-mers. This finding is consistent with observations in prior work such as DNABERT~\citep{DNABert} and Generator~\citep{wu2025generator}, which also identified 6-mer tokenization as optimal for genomic modeling.

\textbf{Model scaling.}
We study scaling behavior by training NT-v2 backbones at 50M, 100M, 250M, and 500M parameters under the same unconditional generation protocol. Figure~\ref{fig:scaling_law} shows that performance is relatively stable from 50M to 250M (SFID 10.92--11.69), indicating that generation quality is not strongly bottlenecked by parameter count in this range. We hypothesize this is due to information saturation relative to the dataset size, where the larger model overfits to high-frequency noise.

\textbf{Sampling strategy.}
We further compare decoding/sampling strategies at inference time, maintaining consistency with previous settings. As shown in Table~\ref{tab:sampling}, MaskGit-style sampling, entropy-based sampling, and Topk-M yield consistently high fidelity (SFID 10.93--11.05), while simple random masking achieves the best SFID (10.92) among the tested strategies. In contrast, P2 produces extremely high diversity but suffers from a catastrophic collapse in fidelity (SFID 3847.58) and exhibits severely distorted GC statistics (12.7 vs.\ $\approx$1.0 for Truth), indicating a failure to preserve basic compositional constraints. Overall, these results suggest that our model is robust to most standard sampling choices, whereas overly aggressive strategies can severely distort the generated distribution. Detailed formulations of these sampling strategies are provided in Appendix~\ref{appendix:sampling}.

\textbf{Denoising Steps and Temperature.}
We also investigated the impact of the number of decoding steps and the sampling temperature during the denoising process on the quality of the generated sequences. Please refer to Appendix~\ref{appendix:steps} and Appendix~\ref{appendix:temperature} for details.

%% file: sections/conclusion.tex
\section{Conclusion}
\label{sec:conclusion}

In this work, we introduced D3LM, which unifies bidirectional representation learning and DNA generation through masked diffusion in discrete token space. By training with variable masking ratios while maintaining bidirectional attention, D3LM addresses fundamental limitations of both BERT-style and autoregressive DNA models. Our work establishes discrete diffusion as a promising paradigm for DNA foundation models.

%% file: sections/related_work.tex
\section{Related work}

\textbf{Genomic language models}. Early genomic models are trained as supervised models, taking DNA sequences as input to predict thousands of genome tracks~\citep{zhou2015predicting,zhou2018deep, enformer, yangspace, avsec2025alphagenome}. Recent genomic language models have drawn inspiration from the success of unsupervised pre-training in natural language processing~\citep{devlin2019bert, brown2020language}. DNABERT~\citep{DNABert} pioneered this approach by adapting BERT's masked language modeling objective to DNA sequences, demonstrating that pre-trained representations transfer effectively to downstream genomic tasks. Subsequent work has continued along two primary directions. BERT-style models~\citep{DNABert2, NT, sanabria2024dna} employ bidirectional masked language modeling with fixed masking ratios, achieving strong performance on understanding tasks but lacking the ability to generate novel sequences for applications such as regulatory element design. Conversely, autoregressive models~\citep{hyenadna, evo} adopt next-token prediction with causal attention, enabling sequence generation but constraining the model to causal left-to-right processing inherited from natural language modeling. 


\textbf{Masked Diffusion Models}. Diffusion models have emerged as powerful generative models in recent years. By defining a forward data corruption process, they learn to reverse the denoising procedure from noisy data~\citep{song2020score, ho2020denoising}. Initially developed for continuous spaces, these models have achieved promising results in continuous data space, such as image generation~\citep{rombach2022high}, audio synthesis~\citep{kong2020diffwave}, human motion generation~\citep{yang2023synthesizing}, and 3D molecular generation~\citep{hoogeboom2022equivariant}. \citet{discdiff} explored encoding discrete DNA sequences into a continuous latent space, followed by performing diffusion in the continuous domain.

Concurrently, researchers began to explore the application of diffusion models in discrete data. D3PM~\citep{d3pm} pioneered discrete diffusion models, defining discrete state transitions as the corruption process for diffusion models and exploring preliminary results in discrete spaces. Masked diffusion models~\citep{nie2025scaling, llada, sahoo2024simple} represent a special class of discrete diffusion models with only two states: preserving the original token or transitioning to a masked state. MDMs can be viewed as BERT~\citep{devlin2019bert} with variable mask ratios, but are fundamentally modeled from a generative perspective, thus possessing generative capabilities. 

A series of theoretical~\citep{sedd, icml_bestpaper} and empirical~\citep{sahoo2024simple, nie2025scaling, llada, dream} works have gradually demonstrated the substantial potential of masked diffusion models for discrete space modeling. Recent efforts in scaling masked diffusion models to billion-scale parameters~\citep{llada, dream} have achieved performance comparable to autoregressive models in text generation.

Notably, MDLM~\citep{sahoo2024simple} introduced a simplified training objective and demonstrated strong performance on language modeling. MDLM also explored applying this framework to DNA sequences, achieving competitive perplexity on the HG38 genome. However, these experiments were conducted on small-scale models (467K parameters) and evaluated only using perplexity-based metrics, which do not capture biological validity. Our concurrent work NT-v3~\citep{boshar2025foundational} fine-tuned their pre-trained model following the MDLM approach for regulatory element generation, but provided only case studies without systematic evaluation. In contrast, our work presents the first comprehensive study of masked diffusion for DNA foundation models. We scale the framework to 500M parameters and conduct systematic evaluation across both understanding tasks and regulatory element generation. Importantly, we introduce biologically meaningful generation metrics including SFID and compositional constraints, and perform extensive ablation studies to understand the role of key components (Section~\ref{sec:ablation}) of masked diffusion models. 

%% file: appendix/metrics.tex
\section{Evaluation Metrics}
\label{appendix:metrics}

We evaluate D3LM on both generative and discriminative tasks using complementary metrics designed to assess different aspects of model performance.

\subsection{Metrics for Understanding Tasks}

For downstream classification tasks, we report the Matthews Correlation Coefficient (MCC), a balanced metric suitable for datasets with class imbalance. For a binary classification task with true positives (TP), true negatives (TN), false positives (FP), and false negatives (FN), MCC is computed as:
\begin{equation}
\text{MCC} = \frac{\text{TP} \times \text{TN} - \text{FP} \times \text{FN}}{\sqrt{(\text{TP}+\text{FP})(\text{TP}+\text{FN})(\text{TN}+\text{FP})(\text{TN}+\text{FN})}}.
\label{eq:mcc}
\end{equation}
MCC ranges from $-1$ (perfect disagreement) to $+1$ (perfect agreement), with $0$ indicating random prediction. Unlike accuracy, MCC accounts for all four confusion matrix categories and remains informative even when classes are imbalanced.

\subsection{Metrics for Generative Tasks}

We assess the quality of generated DNA sequences using four complementary metrics that capture different aspects of generation fidelity.

\textbf{Diversity (Div).} Diversity measures the internal variance within the set of generated sequences, quantifying whether the model produces varied outputs rather than repetitive patterns. Given a set of $N$ generated sequences $\{\mathbf{x}^{(1)}, \ldots, \mathbf{x}^{(N)}\}$, diversity is computed as the average pairwise Levenshtein distance:
\begin{equation}
\text{Div} = \frac{2}{N(N-1)} \sum_{i=1}^{N-1} \sum_{j=i+1}^{N} d_{\text{edit}}(\mathbf{x}^{(i)}, \mathbf{x}^{(j)}),
\label{eq:diversity}
\end{equation}
where $d_{\text{edit}}(\cdot, \cdot)$ denotes the Levenshtein (edit) distance between two sequences. Higher diversity indicates greater variation in the generated distribution.

\textbf{Novelty (Nov).} Novelty quantifies how different generated sequences are from the training corpus, ensuring the model generates novel sequences rather than memorizing training data. For each generated sequence $\mathbf{x}^{(i)}$, we compute its minimum distance to the training set $\mathcal{D}_{\text{train}}$:
\begin{equation}
\text{Nov} = \frac{1}{N} \sum_{i=1}^{N} \min_{\mathbf{x} \in \mathcal{D}_{\text{train}}} d_{\text{edit}}(\mathbf{x}^{(i)}, \mathbf{x}).
\label{eq:novelty}
\end{equation}
Higher novelty indicates that generated sequences are distinct from training examples, demonstrating generalization rather than overfitting.

\textbf{GC Ratio.} The GC ratio measures compositional validity by computing the ratio of Guanine (G) to Cytosine (C) nucleotides. Natural DNA typically maintains approximate strand symmetry (Chargaff's parity rules), with a GC ratio close to $1.0$. For a sequence $\mathbf{x}$:
\begin{equation}
\text{GC Ratio} = \frac{\#\{\text{G in } \mathbf{x}\}}{\#\{\text{C in } \mathbf{x}\}},
\label{eq:gc_ratio}
\end{equation}
where $\#\{\cdot\}$ denotes the count of occurrences. Deviations from $1.0$ indicate violations of basic compositional constraints in DNA.

\textbf{Sei-based Fr\'{e}chet Inception Distance (SFID).} SFID measures the distributional similarity between generated and real sequences in a biologically meaningful feature space. We use Sei~\citep{chen2022sequence}, a deep learning model trained to predict over 20,000 chromatin profiles, to map sequences into a high-dimensional regulatory feature space. Given the feature representations of real sequences $\{\mathbf{f}_r^{(i)}\}$ and generated sequences $\{\mathbf{f}_g^{(i)}\}$, SFID is computed as:
\begin{equation}
\text{SFID} = \|\boldsymbol{\mu}_r - \boldsymbol{\mu}_g\|_2^2 + \text{Tr}\left(\boldsymbol{\Sigma}_r + \boldsymbol{\Sigma}_g - 2(\boldsymbol{\Sigma}_r \boldsymbol{\Sigma}_g)^{1/2}\right),
\label{eq:sfid}
\end{equation}
where $\boldsymbol{\mu}_r, \boldsymbol{\Sigma}_r$ and $\boldsymbol{\mu}_g, \boldsymbol{\Sigma}_g$ are the mean vectors and covariance matrices of real and generated feature distributions, respectively, and $\text{Tr}(\cdot)$ denotes the matrix trace. Lower SFID indicates that generated sequences possess regulatory properties closer to natural DNA, capturing functional similarity beyond sequence-level statistics.

%% file: appendix/sampling.tex
\section{Sampling Strategies}
\label{appendix:sampling}

In the discrete diffusion generative process, the model iteratively recovers masked tokens over $T$ timesteps. At each step $t$, the model predicts the probability distribution $p_\theta(x_0 | x_t)$ for all currently masked positions. Let $\mathcal{M}_t$ denote the set of indices that are masked at step $t$. The core difference between sampling strategies lies in two aspects: (1) how to select the candidate token $\hat{x}_i$ for each masked position $i \in \mathcal{M}_t$, and (2) how to compute a confidence score $c_i$ to decide which tokens to unmask (keep) for the next step $t-1$.

We denote the predicted logits at position $i$ as $z_i \in \mathbb{R}^{V}$, where $V$ is the vocabulary size. The probability distribution is $p_i = \text{Softmax}(z_i / \tau)$, where $\tau$ is the temperature.

\paragraph{1. MaskGit (Confidence-based Sampling)}
Following~\citet{chang2022maskgit}, this strategy selects the most probable token and uses its probability as the confidence measure.
\begin{align*}
    \hat{x}_i &= \arg\max p_i \\
    c_i &= \max(p_i)
\end{align*}
In our implementation, we allow sampling $\hat{x}_i$ from the distribution $p_i$ rather than strictly taking the argmax if $\tau > 0$, but the confidence remains the probability of the selected token. At each step, we unmask a subset of tokens with the highest confidence scores according to a linear schedule.

\paragraph{2. Entropy-based Sampling}
This strategy measures the uncertainty of the model's prediction using Shannon entropy. Lower entropy indicates higher confidence.
\begin{align*}
    \hat{x}_i &\sim p_i \\
    c_i &= - H(p_i) = \sum_{v=1}^{V} p_i^{(v)} \log p_i^{(v)}
\end{align*}
Tokens with lower entropy (higher negative entropy) are prioritized for unmasking. This approach favors positions where the model is certain about the distribution, even if the top-1 probability is not extremely high.

\paragraph{3. Top-k Margin (Topk-M)}
This metric defines confidence as the margin between the most likely token and the second most likely token.
\begin{align*}
    \hat{x}_i &\sim p_i \\
    c_i &= p_i^{(\text{top1})} - p_i^{(\text{top2})}
\end{align*}
where $p_i^{(\text{top1})}$ and $p_i^{(\text{top2})}$ are the largest and second-largest probabilities in $p_i$. A large margin indicates that the model clearly distinguishes the best candidate from alternatives, reducing the risk of sampling ambiguity.

\paragraph{4. Random Sampling}
In this baseline strategy, we disregard confidence scores entirely for the selection of unmasking positions.
\begin{align*}
    \hat{x}_i &\sim p_i \\
    c_i &\sim \text{Uniform}(0, 1)
\end{align*}
Tokens to be unmasked are selected uniformly at random from the set of currently masked tokens $\mathcal{M}_t$. This serves as an ablation to verify the necessity of confidence-based scheduling.

\paragraph{5. P2 (Proposal-based Probability)}
The P2 strategy~\citep{peng2025path} employs a ``re-masking" mechanism distinct from the standard parallel decoding. Instead of monotonically reducing the mask set, P2 allows previously unmasked tokens to be re-masked if their confidence drops.
At step $t$, for \textit{all} positions (both currently masked and unmasked), we compute confidence based on the current prediction (typically using the max probability). We then select the top $k_t$ tokens with the highest confidence to keep unmasked, where $k_t$ follows the diffusion schedule.
\begin{align*}
    \mathcal{U}_{t-1} = \text{TopK}_{i \in \text{All Indices}} (c_i, k_t)
\end{align*}
This allows the model to ``correct" earlier mistakes by re-masking them, but as shown in our experiments (Table~\ref{tab:sampling}), this instability can lead to catastrophic failure in maintaining long-range genomic constraints (e.g., GC ratio) when applied to DNA generation.

%% file: appendix/ablation.tex
\section{Ablation on Tokenization}

\begin{figure*}[h]
\vskip -0.1in
  \centering
  \begin{subfigure}[b]{0.75\textwidth} 
    \centering
    \includegraphics[width=\linewidth]{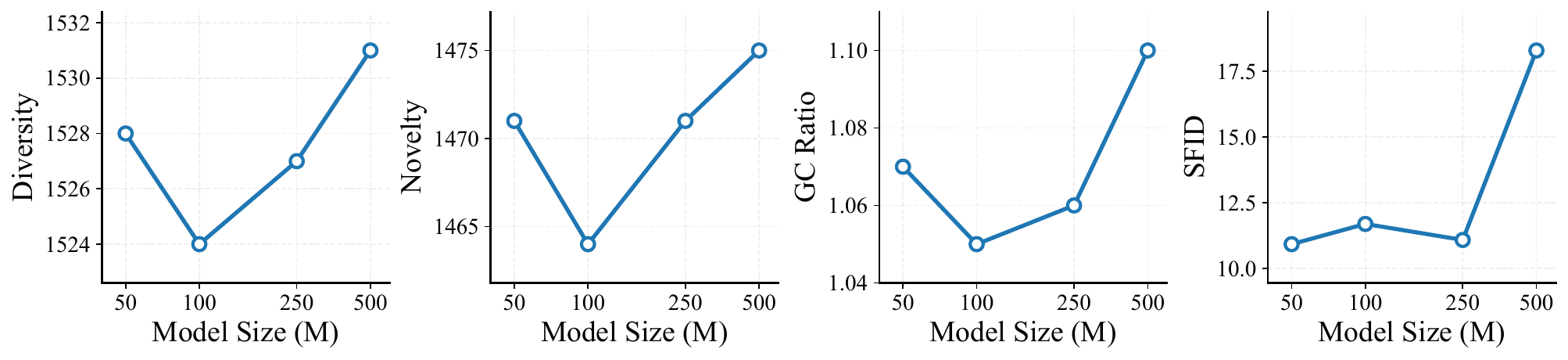}
    \caption{Scaling Law}
    \label{fig:scaling_law}
  \end{subfigure}

  \begin{subfigure}[b]{0.75\textwidth} 
    \centering
    \includegraphics[width=\linewidth]{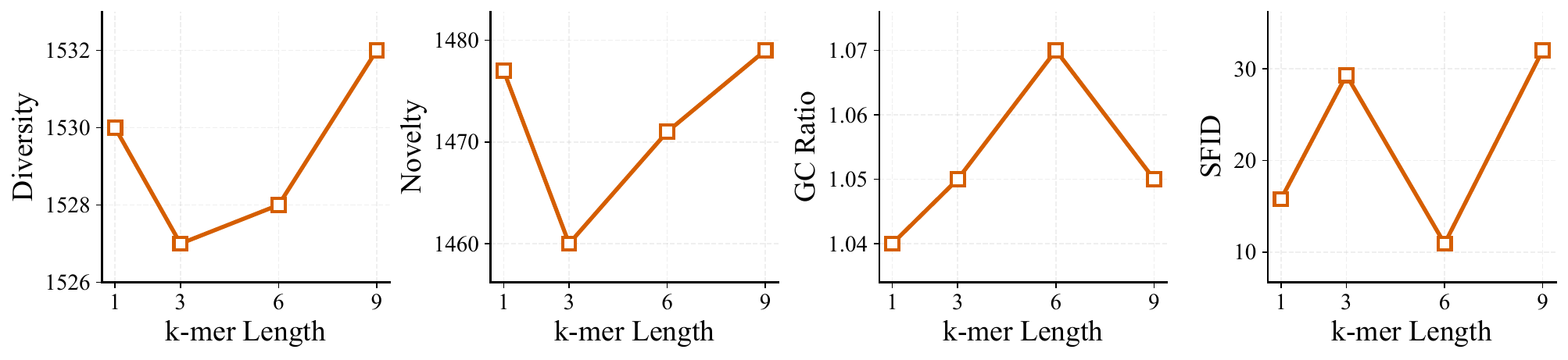}
    \caption{Tokenization}
    \label{fig:tokenization}
  \end{subfigure}
  
  \caption{Analysis of Model Properties. (a) Illustration of the scaling law behavior. (b) Visualization of the tokenization process.}
  \label{fig:model_analysis}
  \vskip -0.2in
\end{figure*}

\section{Ablation on Sampling Strategy}

\begin{table}[h]
  \caption{Abaltion study of sampling strategy.}
  \label{tab:sampling}
  \begin{center}
    \begin{small}
      \begin{sc}
        \begin{tabular}{lcccc}
          \toprule
          strategy & Div ($\uparrow$) & Nov ($\uparrow$) & GC & SFID ($\downarrow$) \\
          \midrule
          Truth & 1529& 1470& 1.06& 7.85 \\
          Random & 1537& 1491& 1.00& 122.97\\
          \midrule
          P2 & \textbf{1662.0} & 1451.0 & 12.7& 3847.58 \\
          MaskGit & 1531.0 & 1469.0 & 1.06& 10.93 \\
          Entropy & 1530.0 & \textbf{1472.0} & 1.07& 10.99 \\
          Topk-M & 1529.0 & 1468.0 & 1.05& 11.05 \\
          Random & 1528& 1471& 1.07& \textbf{10.92}\\
          \bottomrule
        \end{tabular}
      \end{sc}
    \end{small}
  \end{center}
  \vskip -0.1in
\end{table}

\section{Ablation on Denoising Steps}
\label{appendix:steps}
We investigate the impact of the number of denoising steps $T$ on generation quality by varying $T$ from 1 to 400. As shown in Table~\ref{tab:ablation_steps}, increasing the number of steps initially improves biological fidelity significantly, with SFID dropping sharply from 62.40 at $T=1$ to an optimal 10.92 at $T=50$. Interestingly, further increasing the steps to 100 and beyond leads to a slight degradation in fidelity (SFID rises to $\approx 12\text{--}13$), suggesting a potential over-refinement issue where the model may drift from the natural manifold or over-smooth the distribution. Consequently, we select $T=50$ as the default setting, striking the best balance between generation quality and computational efficiency.

\begin{table}[h]
  \caption{Ablation study of denoising steps.}
  \label{tab:ablation_steps}
  \begin{center}
    \begin{small}
      \begin{sc}
        \begin{tabular}{lcccc}
          \toprule
          Steps ($T$) & Div ($\uparrow$) & Nov ($\uparrow$) & GC & SFID ($\downarrow$) \\
          \midrule
          Truth & 1529& 1470& 1.06& 7.85 \\
          Random & 1537& 1491& 1.00& 122.97\\
          \midrule
          1   & \textbf{1536} & \textbf{1488} & 1.04 & 62.40 \\
          5   & 1534 & 1482 & 1.04& 28.96 \\
          10  & 1532 & 1478 & 1.04& 17.90 \\
          50  & 1528& 1471& 1.07& \textbf{10.92}\\
          100 & 1528 & 1467 & 1.06& 12.14 \\
          200 & 1526 & 1468 & 1.07& 13.39 \\
          400 & 1527 & 1467 & 1.06& 12.29 \\
          \bottomrule
        \end{tabular}
      \end{sc}
    \end{small}
  \end{center}
\end{table}

\section{Ablation on Sampling Temperature}
\label{appendix:temperature}

We further analyze the effect of sampling temperature $\tau$ on the trade-off between diversity and fidelity, testing temperatures from 0.9 to 1.2. Table~\ref{tab:ablation_temp} demonstrates that $\tau=1.1$ yields the optimal performance, achieving the lowest SFID (10.92) while maintaining diversity (1528) and GC ratio (1.07) remarkably close to the ground truth (1529 and 1.06, respectively). Lower temperatures (e.g., 0.9, 1.0) reduce diversity and degrade fidelity (SFID 38.50 at $\tau=0.9$), likely due to the model focusing on repetitive, high-probability modes. Conversely, increasing $\tau$ to 1.2 boosts diversity further (1532) but leads to a slight drop in fidelity (SFID 14.74). We therefore adopt $\tau=1.1$ as the default temperature for our main experiments.

\begin{table}[h]
  \caption{Ablation study of sampling temperature.}
  \label{tab:ablation_temp}
  \begin{center}
    \begin{small}
      \begin{sc}
        \begin{tabular}{lcccc}
          \toprule
          Temp ($\tau$) & Div ($\uparrow$) & Nov ($\uparrow$) & GC & SFID ($\downarrow$) \\
          \midrule
          Truth & 1529& 1470& 1.06& 7.85 \\
          Random & 1537& 1491& 1.00& 122.97\\
          \midrule
          0.9 & 1513.0 & 1452.0 & 1.11& 38.50 \\
          1.0 & 1524.0 & 1464.0 & 1.08& 14.31 \\
          1.1 & 1528& 1471& 1.07& \textbf{10.92}\\
          1.2 & \textbf{1532.0} & \textbf{1475.0} & 1.05& 14.74 \\
          \bottomrule
        \end{tabular}
      \end{sc}
    \end{small}
  \end{center}
\end{table}